\documentclass[runningheads]{llncs}
\usepackage[T1]{fontenc}
\usepackage{graphicx}
\usepackage{amsfonts}
\usepackage{array}
\usepackage{hyperref}
\usepackage[usenames]{color}

\usepackage{multirow}
\usepackage{pdflscape}   
\usepackage{array}
\usepackage{rotating}

\usepackage{graphicx}
\usepackage{subcaption}
\usepackage{booktabs}

\usepackage{multirow}

\begin{document}

\title{CrimeNER Demo: Named-Entity Recognition \\in the Crime Domain}

\titlerunning{CrimeNER Demo}

\author{Miguel Lopez-Duran\and
Julian Fierrez\and
Aythami Morales\and
Daniel DeAlcala \and Gonzalo Mancera \and Javier Irigoyen \and
Ruben Tolosana\and
Oscar Delgado\and Francisco Jurado \and
Alvaro Ortigosa}
\authorrunning{M. Lopez-Duran, J. Fierrez, et al.}
\institute{BiometricsAI, Universidad Autónoma de Madrid (UAM), Spain\\
\email{miguel.lopezd@uam.es, julian.fierrez@uam.es}}

\maketitle

\begin{abstract}
We present CrimeNER Demo, an AI-powered platform that enables us to extract general crime-related information from documents and classify them into entity types with two levels of granularity. We provide pretrained NER models on the CrimeNER database, and we give the possibility to users to provide their own annotated data to train models for their own specific cases. This demonstrator aims to promote crime-related NER research and provides a practical tool to automatically extract crime information for researchers and law enforcement agencies. The demonstrator includes: i) Pretrained NER models on the crime domain; ii) Possibility to finetune the models on specific data annotated by the user; and iii) An automatic pipeline to extract and annotate crime entities from documents. The demo platform, a tutorial to run the demo, and a video demonstration are publicly available on GitHub.\footnote{ \url{https://github.com/BiometricsAI/CrimeNER.git}}
\end{abstract}

\section{Introduction}


Law enforcement agencies are increasingly required to extract and process information from crime-related documents. However, the number of documents that need to be processed is increasing rapidly, so manually extracting these data is not feasible. In addition, manually annotating sufficient data to train reliable models from scratch is really time consuming and inefficient \cite{PENA-Layout}. 

Automatic extraction of criminal information from document repositories (including web/html locations) may be viewed as a Named Entity Recognition (NER) task \cite{mancera2025pba}. In forensics, the kind of entity law enforcement agencies are interested in can change from case to case. However, when working in criminal cases, information like who committed or is being accused of a crime, or the agents and agencies involved in a case, are general and typically useful kind of information. NER has a lot of work done in several fields. Initially, it was focused on news and general documents, but it grew rapidly to other areas such as biomedicine~\cite{song2021deep}. In relation to the crime domain, there are works that focus on the extraction of legal entities~\cite{au2022ner} and on cyber threat Intelligence~\cite{wang2022aptner}. However, these works revolve around legal texts or specific types of crime, and are not suitable for general crime extraction for day-to-day criminal cases. Coarse entities are annotated with different colors, while fine entities are annotated as metadata inside the document.

Despite all this work, there is still a huge gap in the NER literature in the crime domain. As an initial step to fill this gap, we present a demonstrator of CrimeNER, a platform to extract crime-related entities from input documents. CrimeNER Demo leverages pretrained NER models in the CrimeNER database (CrimeNER-db)~\cite{lopez2026zero}. We label each detected entity into a predefined set of entity types with two levels of granularity, namely coarse and fine-grained entity types, as done in previous works~\cite{ding2021few}. The coarse entity types defined cover basic information about crimes, e. g., the criminal who committed the offense. Fine-grained entity types are defined to better contextualize each of the coarse entity types extracted from the document, e.g. the typology of the criminal activity mentioned or the number of criminals that committed the crime.

We acknowledge that different law enforcement agencies work with different types of crime, and in languages different from English. For this reason, we also let the users upload their own annotated data to the CrimeNER Demo following our schema and finetune the selected models on their data. In this way, only a small number of annotated samples is enough to extract meaningful crime information, avoiding the need for a large annotated database in each target domain \cite{gonzalo2026lora,miguel26docvqa}

This work presents a demonstrator of the CrimeNER platform and serves researchers and practitioners in the line of Forensic Document Analysis using NER techniques.

\section{CrimeNER: Database}

The CrimeNER database (CrimeNER-db) is a dataset comprising more than 1.5K annotated documents from real-world scenarios, such as terrorist reports or real press notes from the US for the extraction of criminal entities.

The primary goal of CrimeNER-db is to provide researchers with a fine-grained general crime dataset. In order to do that, based on previous work on fine-grained NER databases~\cite{ding2021few}, we define a two-level hierarchy for entity types, coarse and fine-grained, and classify each token as part of a coarse entity with its corresponding fine-grained entity type, or as a non-entity. The coarse and their corresponding fine entity types are as follows:
\begin{itemize}
    \item \textbf{Crime}: Illegal activities performed by an individual or group of people, including terrorist attacks. As fine-grained crime entity types, we consider: Terrorism, Fraud, Illegal Traffic, Theft, Drug-related, Homicide, Sexual crimes, Hate crimes, and Others.
    \item \textbf{Actor}: Individuals or organizations that commit or are accused of committed a crime. The fine actor entity types are: Criminal Person, Criminal Organization, Terrorist Person, and Terrorist Organization.
    \item \textbf{Agent \& Agency}: Individuals and organizations acting against criminal activities or involved in criminal cases. We also consider government officials and bodies as this type of entity. We consider three fine entity subtypes: Law Enforcement, Government, and Legal. \footnote{In the original CrimeNER-db work, the Agent and Agency coarse entities are different types, but we decided to merge them into one, as we found it more informative.}
    \item \textbf{Logistic}: Specific details of the crimes mentioned in the documents that may be useful to agents. The fine logistic entity types are Location, GPE, Date, Weapons \& Explosives, and Money.
\end{itemize}

\section{CrimeNER: System}


\begin{table}[p]
\centering
\setlength{\tabcolsep}{8pt}
\caption{Average Strict and Flexible F1 Scores on coarse and fine entity types after training on CrimeNER-db. Strict evaluation requires the same entity span and type to be correct. Flexible evaluation requires same entity type, and an overlap between the predicted span and the ground truth.}
\label{tab:f1_results}
\begin{tabular}{ccccc}
\toprule
 \textbf{Model} & \multicolumn{2}{c}{\textbf{Coarse}} & \multicolumn{2}{c}{\textbf{Fine}} \\
\cmidrule(lr){2-3} \cmidrule(lr){4-5}
 & \textbf{Strict} & \textbf{Flexible} & \textbf{Strict} & \textbf{Flexible} \\
\midrule
XLM-RoBERTa-Base~\cite{DBLP:journals/corr/abs-1911-02116}          & $\mathbf{0.650}$ & $0.900$ & 
$\mathbf{0.650}$ & $0.879$ \\
DeBERTa-V3-Base~\cite{he2021deberta} & $0.649$ & $\mathbf{0.902}$ & $0.627$ & $\mathbf{0.892}$ \\
RoBERTa-Base~\cite{DBLP:journals/corr/abs-1907-11692}              & $0.620$ & $0.899$ & $0.631$ & $0.882$ \\
AlBERT-Base-V2~\cite{DBLP:journals/corr/abs-1909-11942}            & $0.607$ & $0.881$ & $0.401$ & $0.888$ \\
DistilBERT-Base-Cased~\cite{Sanh2019DistilBERTAD}     & $0.519 $& $0.890$ & $0.643$ & $0.886$ \\
BERT-Base-Cased~\cite{DBLP:journals/corr/abs-1810-04805}           & $0.514$ & $0.839$ & $0.644$ & $0.889$ \\
\bottomrule
\end{tabular}
\vspace{3mm}
\end{table}

\begin{figure}[p]
\centering
\includegraphics[scale=0.085]{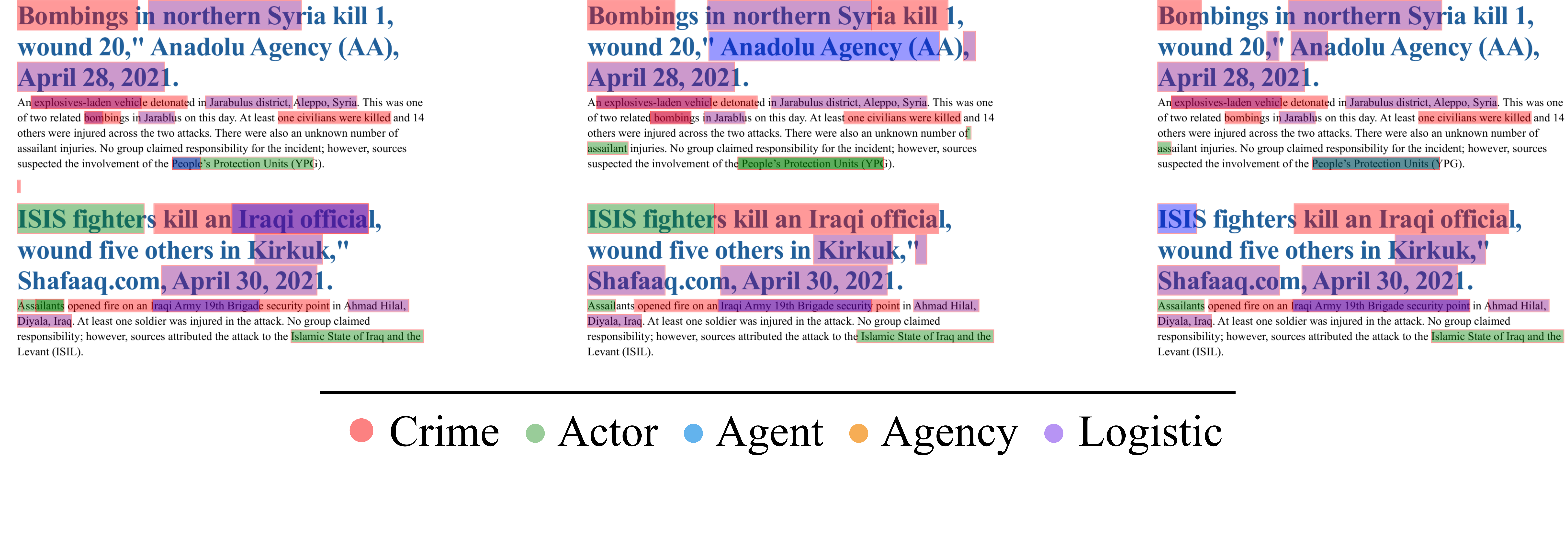}
\vspace{-5mm}
\caption{Example of different annotations for a sample document using the CrimeNER Demo with different model configurations. From left to right, the annotations for coarse entities were made using XLM-RoBERTa-Base~\cite{DBLP:journals/corr/abs-1911-02116}, DeBERTa-V3-Base~\cite{he2021deberta} and DistilBERT-Base-Cased~\cite{Sanh2019DistilBERTAD} and the fine entity annotations were made using DeBERTa-V3-Base, XLM-RoBERTa-Base and AlBERT-Base-V2~\cite{DBLP:journals/corr/abs-1909-11942}. Coarse entities are colored in the document and fine entities are annotated as metadata inside the document.} 
\label{fig::example_2_annotation}
\end{figure}

\begin{figure}[p]
 \centering
 \includegraphics[width=\textwidth]{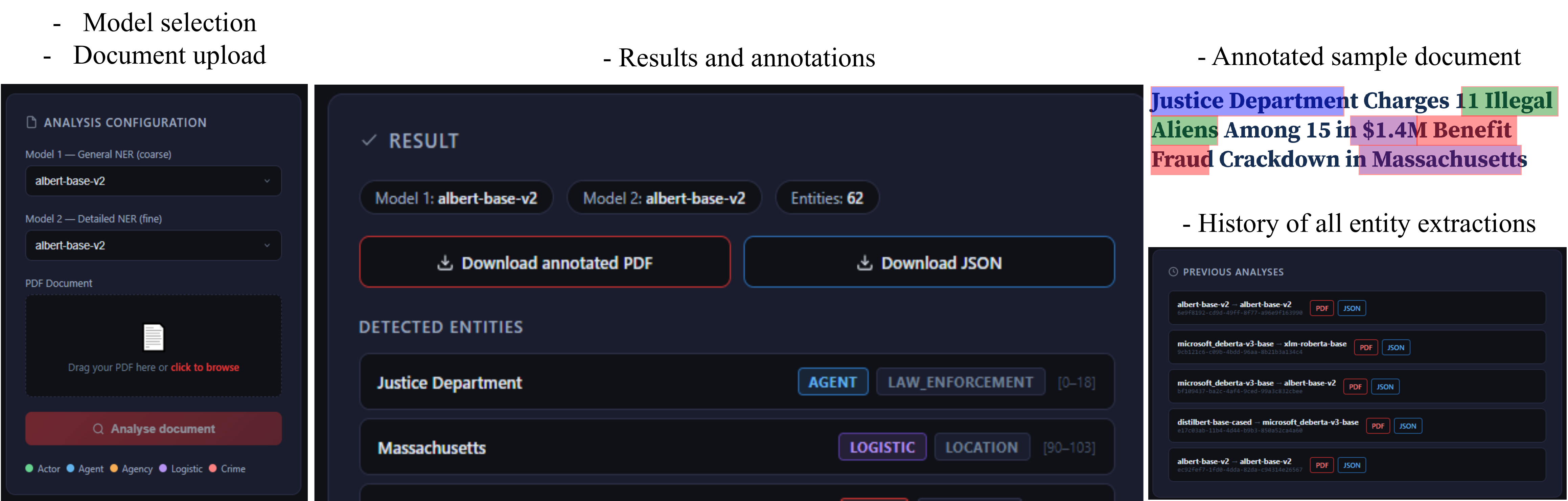} 
  \caption{Overview of the CrimeNER Demo. Documents are processed with the selected model (fine-tuned if specific data is uploaded). Crime-related entities are extracted with two levels of granularity, processed, and shown to the user.}
  \label{fig:crimener system}
\end{figure}

The CrimeNER Demo platform extracts crime related entities from input documents and texts with two levels of granularity. The system comprises 4 processing modules: 1) Document Preprocessing, 2) Specific Fine-tuning (if selected), 3) Coarse and Fine Entity Extraction, and 4) Postprocessing. 

\subsection{CrimeNER Modules}
\subsubsection{Document Preprocessing}

Given an input document or batch of documents, the user is asked to decide whether to process the document using the pretrained NER models or to upload annotated data and finetune them on the data and then process the document. The models were trained in CrimeNER-db by performing a train/val / test split and evaluated on the test split. The results of this training with the selected NER models are shown in Table~\ref{tab:f1_results}.

\subsubsection{Specific Fine-tuning}

Given the wide variety of criminal cases and languages law enforcement agencies may be interested in, we provide the option to further finetune the released criminal NER models on specific target data. This makes our platform even more useful in real-world deployment as it can be adapted to most criminal cases. The only drawback is that the target data need to be formatted as we specify in our demonstrator with the same types of entity. We believe that the types defined on CrimeNER-db are informative enough for most cases. In future work, we will consider new types of entity.

\subsubsection{Coarse and Fine Entity Extraction}

Given the processed documents and the pretrained or finetuned NER models, coarse entities are extracted. After that, on the basis of the extracted coarse entities, we extract the fine entities from them. All this process let us classify each entity into 4 coarse entity types and a total of 21 fine entity types, which makes the extracted information highly detailed and specific for each case. 

\subsubsection{Document Postprocessing}

After the entity extraction, the entity spans are further processed and injected into the input documents. Input documents and texts are returned to the user with the detected crime entity spans highlighted with the detected coarse- and fine-grained entity types detected. A JSON file with the annotation of each document with its extracted entities may also be provided if requested by the user.

\section{CrimeNER: Demonstrator}

This study presents the CrimeNER Demo platform, a platform for extracting general criminal entities from documents. The platform was evaluated using CrimeNER-db, a database consisting of 1.5K real-world documents from the U.S. Department of Justice and other terrorist and crime reports.

The main objective of this work is to promote our platform, where we make CrimeNER Demo available to researchers and practitioners. The CrimeNER Demo allows users to pass input documents or texts and extract the crime-related entities with two levels of granularity within those documents. After processing the documents, users will receive the input documents with the extracted crime entities annotated and highlighted, as shown in Figure~\ref{fig::example_2_annotation}. This platform opens up new opportunities in document analysis and crime detection for researchers and law enforcement agencies.

We provide several pretrained models to extract the crime-related entities. The selection of different models for coarse and fine entities results in different annotations, as shown in Figure~\ref{fig::example_2_annotation}. Qualitative analysis shows that the difference between model annotations is notable, especially the differences in span length for the same entity. There is some misalignment between the annotation and the PDF text, as the positions in the PDF metadata are not exactly aligned with the visual text position in the document.

Figure \ref{fig:crimener system} shows a screenshot of the CrimeNER demonstrator and its main features that we discussed earlier. Initially, users are required to upload one or more documents and select the NER models trained to process the documents. Among these models, there can be the finetuned models on specific data uploaded by the user, for which we provide a tutorial on the CrimeNER demo repository. After model selection, coarse and fine entity types spans are extracted from the documents as we described earlier. Finally, the extracted entities are annotated and highlighted in the documents, and a JSON file with the entity annotations is generated if the user requires it. The CrimeNER platform also stores previous document analysis as part of a history so that users can download documents and annotations already processed again, if necessary.

In future work, we will explore multimodal architectures \cite{2023_SNCS_Human-Centric_Pena}, including the combination of NLP models like those used here and visual models that process text images \cite{miguel26graph,dealcala2026my}. Detecting AI-generated information \cite{dealcala26docai}, 
fakes \cite{MUNOZHARO2026103969}, other types of manipulation \cite{pavel25iccv}, and other general risks \cite{irigoyen26risks} when analyzing document repositories are also key topics on our agenda.

 \subsubsection{\ackname}
 PID2024-160053OB-I00 MICIU/FEDER, Cátedra ENIA (NextGenerationEU PRTR TSI-100927-2023-2), and R\&D Agreement DGGC / UAM/FUAM for Biometrics \& Applied AI. Lopez and Robledo are supported by FPI-UAM-2025, DeAlcala is supported by FPU21/05785, Mancera is supported by FPI-PRE2022-104499, Irigoyen is supported by FPI-PREP2024-003107.

\bibliographystyle{splncs04}
\bibliography{main}

@article{PENA-Layout,
title = {Continuous document layout analysis: Human-in-the-loop {AI} curation, database \& evaluation in the domain of public affairs},
journal = {Inf. Fusion},
volume = {108},
year = {2024},
author = {Alejandro Peña and others},
}

@inproceedings{mancera2025pba,
  title={{{PB}}a-{{LLM}}: Privacy-and bias-aware {{NLP}} using {{N}}amed-{{E}}ntity {{R}}ecognition {{(NER)}}},
  author={Mancera, Gonzalo and Morales, Aythami and Julian Fierrez and others},
  booktitle={ICDAR Workshops},
  year={2025},
}

@inproceedings{gonzalo2026lora,
  title={Auditing Training Data in Domain-adapted {LLMs: LoRA-MINT}},
  author={Gonzalo Mancera and Daniel DeAlcala and Aythami Morales and Julian Fierrez and others},
  booktitle={IEEE COMPSAC},
year={2026},
}

@inproceedings{miguel26docvqa,
  title={Comparative Study of Domain-adapted {VLMs} for General Document Visual Question Answering},
  author={Miguel Lopez-Duran and Elena Marrero and others},
  booktitle={ICDAR Workshops},
year={2026},
}

@inproceedings{dealcala2026my,
  title={Is My Vision-Language Data in Your {AI? Membership Inference Test} {(MINT) Demo 2}},
  author={DeAlcala, Daniel and Gonzalo Mancera and Julian Fierrez and others},
  booktitle={IEEE COMPSAC},
  year={2026}
}

@inproceedings{dealcala26docai,
  title={{DocAI: A} Framework for Generating and Identifying {AI}-edited Documents},
  author={DeAlcala, Daniel and Ching-Yun Ko and Aythami Morales and Julian Fierrez and others},
  booktitle={arXiv preprint},
  year={2026}
}

@inproceedings{pavel25iccv,
  title={{DeepID} Challenge of Detecting Synthetic Manipulations in {ID} Documents},
  author={Pavel Korshunov and others},
  booktitle={IEEE Intl. Conf. on Computer Vision Workshops},
year={2025},
}

@InProceedings{miguel26graph,
author="Lopez-Duran, Miguel
and Julian Fierrez and others",
title="Benchmarking Graph Neural Networks for Document Layout Analysis in Public Affairs",
booktitle="ICDAR Workshops",
year="2025",
}

@ARTICLE { 2023_SNCS_Human-Centric_Pena,
author = {Alejandro Peña and others},
journal = {SN Computer Science},
month = {June},
number = {5},
pages = {434},
title = {Human-Centric Multimodal Machine Learning: Recent Advances and Testbed on {AI}-based Recruitment},
volume = {4},
year = {2023},
}

@article{MUNOZHARO2026103969,
title = {Privacy-aware detection of fake identity documents: methodology, benchmark, and improved algorithms ({FakeIDet2})},
journal = {Inf. Fusion},
volume = {128},
pages = {103969},
year = {2026},
author = {Javier Muñoz and others},
}

@inproceedings{wang2022aptner,
  title={{APTNER}: A specific dataset for {NER} missions in cyber threat intelligence field},
  author={Wang, Xuren and He, Songheng and Xiong, Zihan and Wei, Xinxin and others},
  booktitle={IEEE CSCWD},
  pages={1233--1238},
  year={2022},
}

@inproceedings{ding2021few,
  title={{Few-NERD}: A few-shot named entity recognition dataset},
  author={Ding, Ning and Xu, Guangwei and Chen, Yulin and Wang, Xiaobin and Han, Xu and others},
  booktitle={Proc. ACL},
  pages={3198--3213},
  year={2021}
}

@article{lopez2026zero,
  title={Named-Entity Recognition in the Crime Domain {(CrimeNER): Case} Study and Dataset},
  author={Lopez-Duran, Miguel and Fierrez, Julian and others},
  journal={arXiv:2603.02150},
  year={2026}
}

@article{song2021deep,
  title={Deep learning methods for biomedical named entity recognition: a survey and qualitative comparison},
  author={Song, Bosheng and others},
  journal={Brief. Bioinform.},
  volume={22},
  number={6},
  pages={bbab282},
  year={2021},
  publisher={Oxford University Press}
}

@inproceedings{au2022ner,
  title={{E-NER: An} Annotated Named Entity Recognition Corpus of Legal Text},
  author={Au, Ting Wai Terence and Lampos, Vasileios and Cox, Ingemar},
  booktitle={Proc. of the NLP Workshop},
  pages={246--255},
  year={2022}
}

@inproceedings{DBLP:journals/corr/abs-1911-02116,
  author    = {Alexis Conneau and
               others},
  title     = {Unsupervised Cross-lingual Representation},
  booktitle   = {ACL},
  year      = {2020},
}

@article{DBLP:journals/corr/abs-1907-11692,
  author    = {Yinhan Liu and
               Myle Ott and
               Naman Goyal and
               Jingfei Du and
               Mandar Joshi and
               Danqi Chen and
               Omer Levy and
               others},
  title     = {{RoBERTa: A} Robustly Optimized {BERT} Pretraining Approach},
  journal   = {CoRR},
  volume    = {abs/1907.11692},
  year      = {2019},
  archivePrefix = {arXiv},
  eprint    = {1907.11692},
  bibsource = {dblp computer science bibliography, https://dblp.org}
}

@inproceedings{
he2021deberta,
title={{DeBERTa}: {BERT} with Disentangled Attention},
author={Pengcheng He and others},
booktitle={ICLR},
year={2021},
}

@article{DBLP:journals/corr/abs-1909-11942,
  author    = {Zhenzhong Lan and
               Mingda Chen and
               Sebastian Goodman and
               Kevin Gimpel and
               others},
  title     = {{ALBERT:} {A} Lite {BERT} for Self-supervised Learning of Language
               Representations},
  journal   = {CoRR},
  volume    = {abs/1909.11942},
  year      = {2019},
  archivePrefix = {arXiv},
  eprint    = {1909.11942},
  bibsource = {dblp computer science bibliography, https://dblp.org}
}

@article{Sanh2019DistilBERTAD,
  title={{DistilBERT,} a distilled version of {BERT:} smaller, faster, cheaper and lighter},
  author={Victor Sanh and Lysandre Debut and Julien Chaumond and Thomas Wolf},
  journal={ArXiv},
  year={2019},
  volume={abs/1910.01108}
}

@article{DBLP:journals/corr/abs-1810-04805,
  author    = {Jacob Devlin and
               Ming{-}Wei Chang and
               Kenton Lee and
               Kristina Toutanova},
  title     = {{BERT:} Pre-training of Deep Bidirectional Transformers for Language
               Understanding},
  journal   = {CoRR},
  volume    = {abs/1810.04805},
  year      = {2018},
  archivePrefix = {arXiv},
  eprint    = {1810.04805},
  bibsource = {dblp computer science bibliography, https://dblp.org}
}

@inproceedings{irigoyen26risks,
  title={Overview of Risk Assessment and Management for Intelligent Systems under the {AI Act} and Beyond},
  author={Javier Irigoyen and others},
  booktitle={IEEE ICCST},
  year={2026}
}

\end{document}